\documentclass[letterpaper, 10pt, conference]{ieeeconf}      % Use this line for a4 paper

\usepackage{graphics} 
\usepackage{epsfig} 
\usepackage{mathptmx} 
\usepackage{times} 
\usepackage{amsmath,amssymb,amsfonts} 
\usepackage{cite}
\usepackage{graphicx}
\usepackage{textcomp}
\usepackage{wrapfig}
\usepackage{color}
\usepackage{bm}
\usepackage{multirow}
\usepackage{booktabs}
\usepackage{array}
\usepackage{gensymb}
\usepackage{tabularx}
\usepackage{float}
\usepackage{microtype}
\usepackage{algorithmic}
\usepackage{array}
\usepackage[caption=false,font=normalsize,labelfont=sf,textfont=sf]{subfig}
\usepackage{stfloats}
\usepackage{url}
\usepackage{verbatim}
\IEEEoverridecommandlockouts                              % This command is only needed if 
\begin{document}
\title{\LARGE \bf
Semi-Supervised Disentanglement of\\ Tactile Contact~Geometry from Sliding-Induced Shear
}

\author{Anupam K. Gupta$^{1}$, Alex Church$^{1}$ and Nathan F. Lepora$^{1}$% <-this % stops a space
\thanks{*This work was supported by an award from the Leverhulme Trust on “A biomimetic forebrain for robot touch” (RL-2016-39)}% <-this % stops a space
\thanks{$^{1}$The authors are with the Department of Engineering Mathematics and the Bristol Robotics Laboratory, University of Bristol, UK \mbox{{\tt\small \{anupam.gupta,ac14293,n.lepora\}@bristol.ac.uk}}}%
}

% \author{Anupam K. Gupta and Nathan F. Lepora
% \thanks{This work was supported by an award from the Leverhulme Trust on “A biomimetic forebrain for robot touch” (RL-2016-39)}.
% \thanks{AKG and NFL are with the Department of Engineering Mathematics and the Bristol Robotics Laboratory, University of Bristol, U.K.}
% \thanks{Corresponding author: anupam.gupta@bristol.ac.uk}}

% \markboth{IEEE ROBOTICS AND AUTOMATION LETTERS, Vol. X, No. X, February 2022}
% {GUPTA AND LEPORA: Semi-Supervised Disentanglement of Tactile Contact Geometry from Sliding-Induced Shear}

\maketitle
\thispagestyle{empty}
\pagestyle{empty}

\begin{abstract}
The sense of touch is fundamental to human dexterity. When mimicked in robotic touch, particularly by use of soft optical tactile sensors, it suffers from distortion due to motion-dependent shear. This complicates tactile tasks like shape reconstruction and exploration that require information about contact geometry. In this work, we pursue a semi-supervised approach to remove shear while preserving contact-only information. We validate our approach by showing a match between the model-generated unsheared images with their counterparts from vertically tapping onto the object. The model-generated unsheared images give faithful reconstruction of contact-geometry otherwise masked by shear, along with robust estimation of object pose then used for sliding exploration and full reconstruction of several planar shapes. We show that our semi-supervised approach achieves comparable performance to its fully supervised counterpart across all validation tasks with an order of magnitude less supervision. The semi-supervised method is thus more computational and labeled sample-efficient. We expect it will have broad applicability to wide range of complex tactile exploration and manipulation tasks performed via a shear-sensitive sense of touch.    
\end{abstract}
% \begin{IEEEkeywords}
% Robotic Touch, Disentanglement, Shear, Semi Supervision, Object Reconstruction.
% \end{IEEEkeywords}

\begin{figure*}[t]
  \centering
  \includegraphics[width=.95\textwidth]{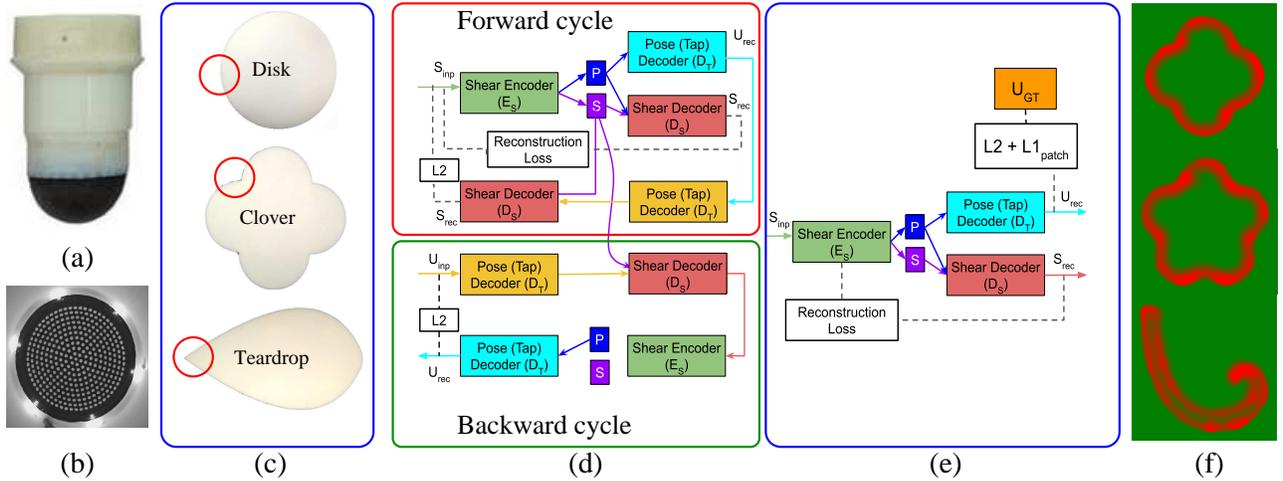}
  \vspace{-.5em}
  \caption{\textbf{Sensor, stimuli, model schematic and object shape reconstructions.} (a) TacTip sensor. (b) TacTip internal sensing surface.  (c) 3D printed tactile stimuli. Red circles highlight approximate location of data collection for training. (d) \& (e) Schematic of the model architecture for self-supervised and supervised training phases respectively. $S_{\rm inp}$ and $U_{\rm inp}$ represent sheared data (distorted by motion-induced shear) and unsheared (vertical tap) input samples. $S_{\rm rec}$ and $U_{\rm rec}$ represent sheared and unsheared reconstructed samples. $U_{\rm GT}$ represents unsheared ground truth samples used for fine-tuning sheared-unsheared mapping in the supervised phase. P \& S are latent representations of pose (contact-only) and shear components of sensor response respectively. $L1_{\rm patch}$ loss is computed between 100 random patches ($20\times20$ pix.) of model-generated unsheared images \textit{$U_{\rm rec}$} and its ground-truth counterparts \textit{$U_{\rm GT}$}, is used to enforce local image compliance. Detailed architectures in Fig.~\ref{fig:base_arch}. (f) Representative examples of full object shape reconstructions. For full set of results, see Fig.~\ref{fig:results}.}
  \vspace{-.5em}
  \label{fig:sensor}
\end{figure*}

\section{Introduction}

The sense of touch is ubiquitous in everyday life. It allows us to interact, explore and manipulate the environment around us using physical contact. The physical contact allows us to measure the environment directly, which is not possible with other sensory modalities like vision. However, this physical contact also results in entangling of relevant information like contact geometry with the manner in which the sensor-stimuli contact is established. For example, when sliding or rubbing fingers across tactile stimuli, soft tactile sensors are inevitably distorted by motion-dependent shear, making their response history-dependent and generalization hard across tactile tasks. This necessitates the removal of motion-induced global shear while preserving the sensor distortions due to the contact-geometry or any other attribute of interest. This issue plagues all shear-sensitive tactile sensors, in particular camera-based soft tactile sensors that use the displacement of markers to encode stimuli attributes~\cite{Nathan_Tactip, Sensor_ChromaTouch, kamiyama_vision-based_2005, Sensor_Andrea}, complicating tactile-dependent tasks like shape modelling and exploration.

Recently, this shear problem was tackled in~\cite{Nathan_CNN_Sliding1, Nathan_CNN_Sliding2} for sliding exploration of complex 2D or 3D objects, using a supervised deep learning model to learn mappings between shear-distorted data and the target pose. However, this approach to learn insensitivity to shear requires training separate models for each tactile task, such as contact reconstruction or shape exploration. A study by the authors of this paper instead used a supervised deep learning model to remove shear distortion from the original sheared tactile image so that quantities of interest (e.g. pose) can be predicted from unsheared tactile data in a task-agnostic manner, which is more computational and data efficient~\cite{Anupam_Disentangle_Super}. However, because these approaches are supervised, they suffer from costly burden of requiring annotated data for learning. This limitation can be overcome by the use of self- and semi-supervised approaches that can learn rich representations from uncurated/unlabelled datasets, and are thus more \textit{labeled sample-efficient}. This mode of learning is closer to human learning, as we likewise learn with only small amounts of labelled data and a vast resource of self-supervised experiences. 

In this work, we propose a semi-supervised learning approach to remove shear from shear-distorted tactile images that uses only a small subset of annotated data (10\%) to achieve comparable performance to previous fully-supervised approaches~\cite{Anupam_Disentangle_Super}. Self supervision with tactile data differs from computer vision because it lacks attributes like colour, texture and the large-scale structure of natural image data that have been exploited in computer vision to generate supervisory signals without external supervision ~\cite{Afros_SplitBrainAE, Noroozi_Jigsaw, Afros_Inpainting}. Moreover, the present work relies only on tactile data without recourse to techniques that correlate data across multiple sensory modalities to generate training signals without external supervision~\cite{Owens_AudioVisual_SemiSuper, Raia_Touch_SemiSuper, Lee_Touch_SemiSuper} (see Background). In the presence of these limitations, we achieved self-supervision by first disentangling sliding data into its contact-only and shear components, followed by recombining the shear component with paired and novel canonical (non-sheared or vertical tap) data to generate training signals without external supervision. %(see Methods). 

Our experiments demonstrated that the proposed model could successfully recover contact geometry from sheared images, allow continuous shape exploration and also allow object reconstruction across multiple planar objects. We also found that our proposed semi-supervised model achieved near identical performance to its fully supervised counterpart~\cite{Anupam_Disentangle_Super} while using only 10\% of labelled samples for training.% in comparison to~\cite{Anupam_Disentangle_Super}.

\section{Background}
Soft optical tactile sensors like the TacTip~\cite{Nathan_Tactip} and other marker-based sensors~\cite{Sensor_ChromaTouch,Sensor_Andrea} are vulnerable to sliding-induced global shear that distorts the geometry of contact. Local shear caused by the stimulus imprint on the sensor, represented in the lateral displacement of the markers, is superimposed upon a global shear caused by contact motion such as sliding. This distortion, by masking contact geometry, complicates challenging tactile tasks like shape reconstruction and tactile exploration, which require spatial information about the local geometry of an object free from motion-induced distortion.

Recent works that use TacTip sensor (also used here) addressed this shear problem by using a shear-insensitive pose-prediction deep network to slide around complex planar shapes~\cite{Nathan_CNN_Sliding1} and complex 3D objects~\cite{Nathan_CNN_Sliding2,Nathan_CNN_Sliding3}. Shear-insensitive training was achieved by mapping the shear-distorted samples directly to target pose outputs. While effective, this approach necessitated separate training for each new tactile task, for example surface- and contour-following~\cite{Nathan_CNN_Sliding3}, greatly complicating the hyper-parameter tuning for accurate network performance~\cite{Nathan_CNN_Sliding2}, and could not be applied to shape reconstruction. To overcome these drawbacks, the authors of the present paper instead focused on learning a mapping between shear-distorted tactile images and their unsheared (vertical discrete tap) counterparts with a fully-supervised convolutional neural network~\cite{Anupam_Disentangle_Super}. Once learned, this mapping can be reused to remove shear for any number of downstream tasks, making this approach computational and data efficient. However, the supervised nature of that approach necessitated complex labelled-data collection. Here, we improve upon that work by proposing a semi-supervised model that achieves comparable performance to the previous supervised baseline, by training first in self-supervised phase followed by a brief supervised phase (with a tenth of the labelled data from~\cite{Anupam_Disentangle_Super}). 

Tactile data differs from natural images in being low-dimensional due to absence of attributes like color and texture, alongside lacking the large-scale global structure of natural image data. This is primarily due to the localised nature of tactile sensing, unlike vision or audition which are global sensing modalities over a scene. This increases the complexity of generating a supervisory signal in the absence of external supervision; for example, in computer vision, supervisory signals in prior studies have been generated by: 1) learning to predict withheld data dimensions at the output, such as in~\cite{Afros_SplitBrainAE} two sub-networks are trained that are each tasked with predicting one subset of the data channels (color or lightness channel) from another (lightness or color channel); 2) image completion conditioned on the immediate surroundings, as in ~\cite{Afros_Inpainting} in which the input image has a missing part that needs to be predicted at the output based on the surroundings; and 3) predicting the correct arrangement of image subsections, as in ~\cite{Noroozi_Jigsaw} in which the input image is first split into $N$ rectangular tiles whose locations are shuffled to break spatial structure, and the network is then tasked to predict the original image from a shuffled input image. These methods cannot be readily extended to tactile data due to its low-dimensional structure and lack of global structure as discussed previously. 

Another approach to generate supervisory signals in absence of external supervision is by correlating data from multiple sensory modalities like audio and vision~\cite{Owens_AudioVisual_SemiSuper} or touch and vision~\cite{Raia_Touch_SemiSuper, Lee_Touch_SemiSuper}. In this work, however we rely on the information encoded by the tactile sensor alone.% that excludes above approaches. 

To generate a supervisory signal, like in ~\cite{Anupam_Disentangle_Super}, the proposed model learns to disentangle the sensor response components due to contact geometry and motion-induced shear. Our approach, like~\cite{Anupam_Disentangle_Super}, is inspired from similar works in computer vision where disentanglement of object attributes and factors of variation are considered important for learning robust representation that improve generalizability on out-of-distribution (OOD) samples or novel scenarios~\cite{bengio_disentanglement}. Several works in computer vision learn disentangled representations via varitional autoencoders (VAE)~\cite{betaVAE, factorVAE} and generative adversarial networks (GAN)~\cite{disentagleGANSS, InfoGAN, IBGAN} for downstream applications such as style transfer between images\cite{styletransfer, disentagleGANSS}. Our approach is akin to style transfer where the input image is decomposed into its content (geometry) and style (color, intensity, texture) components. Once the content and style components are disentangled, they can be used to transfer style to novel content. In our work, content is akin to sensor response due to contact geometry and style to response due to motion-induced shear. We are aware of no other prior work exploring semi-supervised learning of disentangled representations for robust representation learning in tactile sensing.

\begin{figure*}[t]
  \centering
  \includegraphics[width=0.95\textwidth]{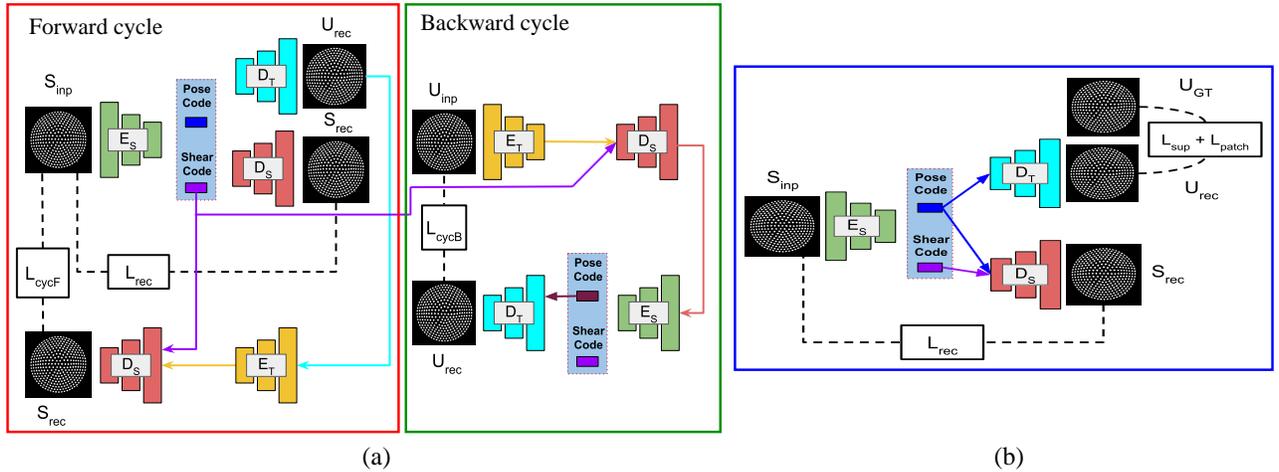}
  \vspace{-.5em}
  \caption{\textbf{Network architecture.} (a) Self-supervised training: The sheared ($S_{\rm inp}$) -- unsheared ($U_{\rm rec}$), unsheared ($U_{\rm inp}$) -- sheared ($S_{\rm rec}$) and sheared ($S_{\rm inp}$) -- sheared ($S_{\rm rec}$) mappings $M$, $K$ and $N$ are learned via cycle consistency loss ($L_{\rm cyc}$)~\cite{Zhu_CycleConsistLoss2017} and reconstruction loss ($L_{\rm rec}$). $L_{\rm cycF}$ and $L_{\rm cycB}$ are the components of ($L_{\rm cyc}$) corresponding to the forward and backward cycle respectively. (b) Supervised training: The mappings $M$ and $N$ previously learned in (a) are further fine-tuned via $L_{sup}$ loss (M), $L1_{\rm patch}$ (M) and reconstructions loss $L_{\rm rec}$ (N). $L1_{\rm patch}$ loss, computed between 100 random patches ($20\times20$ pix.) of model-generated unsheared images \textit{$U_{\rm rec}$} and its canonical (non-sheared) counterparts \textit{$U_{\rm GT}$}, is used to enforce local image compliance.} 
  \vspace{-1em}
  \label{fig:base_arch}
\end{figure*}

\section{Methods}
\subsection{Experimental Setup}
%The variability due to manual assembly of the sensor introduces minor variations in the sensor response of different instances of otherwise identical sensors. 
The setup used in this work is similar to that of~\cite{Anupam_Disentangle_Super}, where a soft biomimetic optical tactile sensor -- the TacTip~\cite{Nathan_Tactip} -- is mounted on an industrial robot arm (ABB IRB120) as an end effector. The tactile sensor encodes tactile information in the shear displacement of marker pins on the inner side of the sensing surface caused by the soft sensor skin deformation. The morphology of this TacTip consisted of a 3D-printed soft rubber-like hemispherical dome (40\,mm dia., TangoBlack+) with its inner surface covered by 331 hard white tip pins arranged in a concentric circular grid fashion (Figs~\ref{fig:sensor}a,b). The soft dome was filled with an optically-clear silicon gel to mimic the compliance of a human fingertip. An RGB camera (ELP 1080p module) captures the inner surface of the sensor dome. For more details, we refer to Ref.~\cite{Nathan_Tactip}. 

For test stimuli, we used seven distinct planar shapes with various morphologies, including two acrylic spiral shapes that differ in frictional properties to the other five 3D-printed ABS plastic shapes (Fig.~\ref{fig:results}, second row). For model training we used three shapes: the circular disk, clover and teardrop shown in Fig.~\ref{fig:sensor} (c) (red circles show approximate location of data collection). All shapes were securely fastened to the workspace to prevent accidental motion during experiments.

\subsection{Data Collection}
This work proposes a semi-supervised deep learning model, as an alternative to the fully-supervised model proposed in~\cite{Anupam_Disentangle_Super}, to remove motion-induced shear distortion from tactile images. This motion-induced shear is governed by the manner of contact between the sensor and object, distorting the sensor response induced by the stimulus geometry during contact.

Our semi-supervised model is trained in two phases: self-supervised and supervised. To ease data collection for these model training phases, we reused the paired data (tap or canonical and sheared data with same relative pose) previously collected in~\cite{Anupam_Disentangle_Super} to train the fully-supervised model. We generate unpaired data, used for self-supervised training phase,  by randomizing the pose pairings (every epoch) between the canonical (non-sheared) tactile images \textit{$U_{\rm inp}$} (collected by vertically tapping on the stimulus) and sheared tactile images \textit{$S_{\rm inp}$} with random global shear (collected by sliding across the stimulus),  for all but a small subset (10\%) of paired data held back for supervised training phase.

% Our semi-supervised model is trained in two phases. In the first phase, the model is trained in a self-supervised fashion on a set of unpaired canonical (non-sheared) tactile images \textit{$UI_{\rm inp}$} (collected by vertically tapping on the stimulus) and a set of (sheared) tactile images \textit{$SI_{\rm inp}$} with random global shear (collected by sliding across the stimulus). This is then followed by a second training phase in which the model is trained on a subset (10\%) of paired canonical and sheared data that was previously used to train the fully-supervised counterpart~\cite{Anupam_Disentangle_Super}. To ease data collection, we reused the paired data collected in~\cite{Anupam_Disentangle_Super} to train the fully-supervised model. However, we randomized the pairings (every epoch) between the canonical and sheared data for all but a small subset (10\%) of data held back for a brief supervised training phase.  

To collect non-sheared canonical tap data, the sensor is brought in the contact with the stimuli vertically to minimise global shear. For sliding data, the sensor is brought to the target pose while in contact with the object from a random offset location along either the $x$- or $y$-axis or both lateral directions simultaneously. To obtain paired data, the tap and sliding data are collected for same relative target poses between the sensor and stimulus (Fig.~\ref{fig:sensor}c). In total, a set of 200 random poses were collected, with the sensor location relative to the initial contact location randomly sampled from a uniform distribution spanning a range $[-5,5]$\,mm in the two lateral directions ($x$- and $y$-axes), $[-45, 45]$\,deg in yaw ($\theta {\bm n}_{z}$) and $[-6,-1]$\,mm in depth ($z$-axis). Finally, the tactile images captured are cropped and subsampled to a $256\times256$-pixel region and binarized to minimise any effect of lighting changes inside the sensor. 

\subsubsection{\textbf{Canonical Data}}
The canonical tapping dataset had 30,000 samples in total: 50 instances of each of the 200 random poses for the three stimuli shapes used for training (Fig.~\ref{fig:sensor} (c)). The instances were generated by varying the indentation depth ($z$-axis) randomly between $[-1, 1]$\, mm from a set indentation depth. This dataset was used as the target (reference) to which the sliding data should be restored.  

\subsubsection{\textbf{Sheared Data}}
The sheared data had 90,000 samples in total: 150 instances for each of the 200 random poses for each of the three stimuli shapes (Fig.~\ref{fig:sensor} (c)). To generate instances, the sensor was first brought in contact with the stimulus at a lateral offset location sampled randomly between $[-2.5, 2.5]$\,mm laterally (along the $x$- or $y$-axes or both), then sliding to the target pose. 

\subsubsection{\textbf{Training and Test Data}}
The canonical and sheared data were first partitioned randomly into training, validation and test sets in the ratio 60:20:20. Thus, for each stimulus shape (Fig.~\ref{fig:sensor}), 120 poses were aligned to training set and 40 poses each to validation and test sets.

The paired training set (canonical-sheared) was further randomly split in the ratio 90:10. The larger subset (90\%) was shuffled to generate random paired data for the first self-supervised phase of model training. The shuffling was repeated before every training epoch to increase variability in training data. The other subset (10\%) kept the pairing between sheared data and a canonical instance of the same pose, for the supervised phase of training.

\subsection{Formulation}
Our goal is to learn mapping functions between two domains, tap (T) and sheared (S), given training samples comprising canonical (tap) data $\{x_{t}\}_{i=1}^{N}$ where $x_{t}\in X_{t}$ and shear-transformed training samples $\{x_{s}\}_{i=1}^{N}$ where $x_{s}\in X_{s}$. Our model includes three mappings, $M: X_{s} \rightarrow X_{t}$, $N: X_{s} \rightarrow X_{s}$ and $K: X_{t} \rightarrow X_{s}$. Our objective has four loss terms: \textit{reconstruction loss} to match the input \textit{$S_{\rm inp}$} and reconstructed sheared samples \textit{$S_{\rm rec}$} to learn the mapping $N$, an \textit{L2 loss} to match the reconstructed \textit{$U_{\rm rec}$} and canonical samples \textit{$U_{\rm GT}$} in the supervised training phase to fine tune the learned mapping~$M$, an \textit{L1 patch loss} to enforce image similarity locally between the reconstructed \textit{$U_{\rm rec}$} and canonical samples \textit{$U_{\rm GT}$} in the supervised training phase to fine tune the learned mapping~$M$ and \textit{\rm cycle consistency loss} to learn mappings $M$ and $K$ and prevent them from contradicting each other. %The detailed mathematical formulation is given below.

The encoder \textit{$E_{S}$} (Fig. 2) disentangles the latent representation ($\mathbf{z \in Z}$) of input sheared samples \textit{$S_{\rm inp}$} into a pose code ($\mathbf{z_{t} \in Z_{t}}$) and shear code ($\mathbf{z_{s} \in Z_{s}}$). This disentanglement of sensor response not only aids in learning of robust mapping $M$ as shown in ~\cite{Anupam_Disentangle_Super}, but also allows recombination of unpaired shear and pose codes to synthesize novel sheared samples necessary for generating training signals via \textit{\rm cycle consistency loss}~\cite{Zhu_CycleConsistLoss2017} without external supervision. 

\paragraph{\rm cycle Consistency Loss ($L_{\rm cyc}$)}
We apply the cycle consistency loss $\mathbb{L}_{\rm cyc}$ to train the model (Fig.~\ref{fig:base_arch}) in the self-supervised phase of training. This loss aids learning of mappings $M$ and $K$, and prevents them contradicting each other. Enforcing cycle-consistency reduces the space of possible mapping functions that can be learned. For each image $x_{s}$ from the domain $S$, the forward cycle should be able to bring the input image $x_{s}$ back to original image i.e. $x_{s} \rightarrow M(x_{s}) \rightarrow K(M(x_{s})) \approx x_{s}$. Similarly, the backward cycle should satisfy  $x_{t} \rightarrow K(x_{t}) \rightarrow M(K(x_{t})) \approx x_{t}$ for each $x_{t}\in T$. This behaviour is enforced via the \textit{\rm cycle consistency loss}:
\vspace{-0.25em}
\begin{equation} 
\begin{split}
\mathbb{L}_{\rm cyc}(M, K) &= \mathbb{E}_{x_{s}\sim X_{s}}  [\|x_{s} - K(M(x_{s}))\|_{2}] \\ &+ \mathbb{E}_{x_{t}\sim X_{t}}  [\|x_{t} - M(K(x_{t})\|_{2}].
\end{split}
\end{equation}

\paragraph{Reconstruction Loss ($L_{\rm rec}$)}
We apply the reconstruction loss $\mathbb{L}_{\rm rec}$ to learn the mapping $N: X_{s} \rightarrow X_{s}$ between input sheared samples \textit{$S_{\rm inp}$} and reconstructed output sheared samples \textit{$S_{\rm rec}$} through a decoder \textit{$D_{S}$}, in both self-supervised and supervised phases of training. This loss aids the disentanglement of sheared-input samples into pose and shear codes: 
\vspace{-0.25em}
\begin{equation}
    \mathbb{L}_{\rm rec}(N) = \mathbb{E}_{x_{s}\sim X_{s}}
    [\|x_{s} - N(x_{s})\|_{2}].
\end{equation}

\paragraph{$L_{sup}$~Loss}
We apply an L2 loss ($\mathbb{L}_{\rm sup}$) to fine tune the mapping $M: X_{s} \rightarrow X_{t}$ between input sheared samples \textit{$S_{\rm inp}$} and model-generated unsheared images \textit{$U_{\rm rec} (M(x_s))$}.
\vspace{-0.25em}
\begin{equation}
    \mathbb{L}_{\rm sup}(M) = \mathbb{E}_{x_{s}\sim X_{s},  x_{t}\sim X_{t}} [\|x_{t} - M(x_{s})\|_{2}].
\end{equation}

\paragraph{$L1_{\rm patch}$~Loss}
We apply $L1_{\rm patch}$ loss $\mathbb{L}_{\rm patch}$ to fine tune the mapping $M: X_{s} \rightarrow X_{t}$. Unlike $\mathbb{L}_{\rm sup}$, this loss helps enforce image similarity locally, because the change in tactile sensor response with contact conditions is predominantly local in nature. The loss was computed between 100 random  crops of size $20\times20$ of model-generated unsheared images \textit{$U_{\rm rec} (M(x_s))$} and its canonical (tap) counterparts \textit{$U_{GT} (x_t)$ }.
\vspace{-0.25em}
\begin{equation} 
\begin{split}
\mathbb{L}_{\rm patch}(M) = \mathbb{E}_{x_{s}\sim X_{s},  x_{t}\sim X_{t}}  [\|&{\rm crops}(x_{t}) -  {\rm crops}(M(x_{s}))\|_{1}].
\end{split}
\end{equation}

\paragraph{Full Objective}
Our full objectives for Encoders \textit{$E_{S}$}, \textit{$E_{T}$} and Decoders \textit{$D_{S}$}, \textit{$D_{T}$} are:
\vspace{-0.25em}
\begin{eqnarray}
        \mathbb{L}_{E_{S}} = \mathbb{L}_{\rm cyc}(M, K) + \mathbb{L}_{\rm rec}(N) + \mathbb{L}_{\rm sup}(M) + \lambda\cdot\mathbb{L}_{\rm patch}(M) \\
        \mathbb{L}_{E_{T}} = \mathbb{L}_{\rm cyc}(M, K) \\
        \mathbb{L}_{D_{T}} = \mathbb{L}_{\rm cyc}(M, K) + \mathbb{L}_{\rm sup}(M) + \lambda\cdot \mathbb{L}_{\rm patch}(M)\\
        \mathbb{L}_{D_{S}} = \mathbb{L}_{\rm rec}(N)
\end{eqnarray}
where $\lambda=0.1$ is the relative loss scale factor.

The fully-supervised counterpart~\cite{Anupam_Disentangle_Super} of the above model is akin in architecture and training methodology to supervised phase of the proposed model except in the amount of annotated data (10$\times$) used for training.

% \textcolor{blue}{The fully-supervised counterpart~\cite{Anupam_Disentangle_Super} of the above model is trained only in supervised fashion with architecture and training details similar to the supervised phase of the proposed model.} %\textcolor{red}{check the training details}}  

\subsection{Implementation}
\subsubsection{Network Architecture}
\paragraph*{\textbf{Shear Encoder}} The shear encoder \textit{$E_{S}$} compressed the sheared input \textit{$S_{\rm inp}$} ($256\times256$-pixel tactile image) using four convolutional (Conv) layers, each followed by batch normalization (BN) and rectified linear unit (ReLU) activation layers respectively (architecture in Fig.~\ref{fig:base_arch}). The convolution layers had 32, 64, 64 and 128 filters respectively. The output of the last convolutional layer was passed to two additional Conv layers whose outputs represent the two latent codes: Pose ($8\times8\times64$) and Shear ($8\times8\times64$).

\paragraph*{\textbf{Pose (Tap) Decoder}} The pose decoder \textit{$D_{T}$} upsampled the pose code to reconstruct unsheared output \textit{$U_{\rm rec}$} ($256\times256$-pixel tactile image). It had five transposed convolutional layers (Tconv), each followed by a BN and ReLU activation layers respectively except the output layer which used sigmoid activation function. The five transposed convolution layers had 64, 64, 32, 32 and 1 filter respectively.

\paragraph*{\textbf{Shear Decoder}} The shear decoder \textit{$D_{S}$} upsampled the pose and shear code to reconstruct sheared input \textit{$S_{\rm rec}$} ($256\times256$-pixel tactile image). It had the same overall architecture as the pose decoder \textit{$D_{T}$} except the number of filters used in Tconv layers. The five transposed convolution layers had 128, 64, 64, 32 and 1 filter respectively.

\paragraph*{\textbf{Pose (Tap) Encoder}} The pose encoder \textit{$E_{T}$} compressed the unsheared output of \textit{$D_{T}$} to $8\times8\times64$ using five Conv layers each followed by a BN and ReLU activation layer. The five convolution layers had 32, 32, 64, 64 and 64 filters.

All encoders and decoders used a kernel of $4\times4$ with stride of 2. For detailed architecture, see Fig.~\ref{fig:base_arch}.

\subsubsection{Training Details}
The input and output tactile data (binary $256\times256$-pixel images) were scaled to the range $[0, 1]$. The network weights were initialized from a zero-centered normal distribution with standard deviation 0.02.  

For training, a batch size of 32 was used. All network layers used L1/L2 regularization along with random image shifts of 1-2\% to prevent overfitting. We used ADAM optimizer~\cite{Kingma_AdamOpt} with $\beta_1$ = 0.5, $\beta_2$ = 0.999 and learning rate of $5\times10^{-5}$ for self-supervised (SS) and $2.5\times10^{-5}$ for supervised (S) training phases respectively. The model was trained for 50 epochs in the SS phase and for another 50 epochs in the S phase. We used learning rate scheduling for S training phase with training rate reduced to one-tenth after 25 epochs. The training and optimization of the networks was implemented in the Tensorflow 2.0 library on a NVIDIA GTX 1660 (6~GB memory) hosted on a Ubuntu machine.

\section{Experimental Results}
\subsection{Disentanglement of Latent Representations}

We did an ablation study to verify the successful disentanglement of sensor response components due to the geometry of contact (pose code) and that induced by motion-induced shear (shear code). To do this, we reconstructed unsheared samples ($U_{\rm rec}$) from the shear code instead of the pose code. This resulted in a $\sim$5-fold increase in mean-squared error (MSE) between test set model-generated unsheared images \textit{$U_{\rm rec}$} and their canonical counterparts \textit{$U_{\rm GT}$} (0.11 from 0.024) as well as $>70$\% reduction in image similarity (SSIM) (18\% from 92.5\%), indicating the two images are structurally very dissimilar~\cite{ssim}. This shows that the appropriate information to reconstruct unsheared samples ($U_{\rm rec}$) resides in the pose code. Similarly, we tested the reconstruction of sheared samples ($S_{\rm rec}$) from only the pose code, which resulted in a 60-fold increase in MSE error between the input \textit{$S_{\rm inp}$} and the reconstructed \textit{$S_{\rm rec}$} sheared samples, and $>70$\% reduction in SSIM between \textit{$S_{\rm inp}$} and \textit{$S_{\rm rec}$} (18\% from 93\%), showing that the both shear and pose code are jointly required for reconstruction of $S_{\rm rec}$. These two experiments jointly demonstrate the successful disentanglement of the sensor response by our model (Fig.~\ref{fig:base_arch}). 

% We did an ablation study to verify the successful disentanglement of the latent representation into pose (non-sheared) and shear codes respectively. To do this, we first reconstructed the unsheared sample using shear code instead of pose code. This resulted in an approximately 5-fold drop in mean-squared error (MSE) between test set model-generated unsheared images \textit{$U_{\rm rec}$} and its canonical counterparts \textit{$U_{\rm GT}$} (0.11 from 0.024). Similarly, a significant drop from 92.5\% to 18\% was observed in image similarity index measure (SSIM). We then reconstructed the sheared input \textit{$S_{\rm rec}$} only using the pose code to result in a 60-fold drop in MSE error between the input \textit{$S_{\rm inp}$} and reconstructed \textit{$S_{\rm rec}$} sheared samples, and over a 70\% drop in image similarity between \textit{$S_{\rm inp}$}and \textit{$S_{\rm rec}$} (18\% from 93\%). These studies demonstrate the successful disentanglement of the latent representation by our model (Fig.~\ref{fig:base_arch}).

\subsection{Motion-induced Shear Removal from Tactile Images}
First, we used multi-scale SSIM~\cite{ssim} to test the effectiveness of our approach in removing motion-induced global shear. To do so, we computed the average image similarity between the model-generated unsheared images \textit{$U_{\rm rec}$} and their test set sheared \textit{$S_{\rm inp}$} and canonical (non-sheared) \textit{$U_{\rm GT}$} counterparts. On comparison, we found a closer match between \textit{$U_{\rm rec}$} and \textit{$U_{\rm GT}$} (92.5\% with 10\% supervision), dropping to 74\% in absence of any supervision) than \textit{$U_{\rm rec}$} and \textit{$S_{\rm inp}$} (32\%). The corresponding values for fully supervised baseline~\cite{Anupam_Disentangle_Super} were 93\% and 32\% respectively. Thus, our model not only successfully removes global shear from tactile images but is also economical with labelled data (using just 10\% of the previous supervised method~\cite{Anupam_Disentangle_Super}). Degraded performance was seen both in the supervised baseline~\cite{Anupam_Disentangle_Super} and semi-supervised approach proposed in this work on reducing the labelled data used for training. For example, a 20\% reduction in labelled training data lead to a drop of approximately 2\% (from 93\%) and 0.5\% (from 92.5\%) in image similarity between the model-generated unsheared images and their test set canonical (non-sheared counterparts) for supervised baseline~\cite{Anupam_Disentangle_Super} and proposed semi-supervised approaches respectively.

We also found that omitting $L1_{\rm patch}$ loss leads to a 1.5\% drop in image similarity between test set canonical images \textit{$U_{\rm GT}$} and their model-generated unsheared counterparts \textit{$U_{\rm rec}$} (91.5\% from 93\%)at 10\% supervision.

% \textcolor{blue}{We also used average image similarity  between \textit{$U_{\rm rec}$} and \textit{$U_{\rm GT}$} to investigate the impact of $L1_{\rm patch}$ loss on the performance of the model. Omitting $L1_{\rm patch}$ loss from the objective function reduced the average image similarity from 92.5\% to $\sim$91\% at 10\% supervision.}

\begin{table}[h!]
\centering
\caption{\textbf{Mean Absolute Error (MAE) of Pose Prediction}}
\label{tab:poseerr}
\resizebox{0.48\textwidth}{!}{%
\begin{tabular}{|c|c|c|c|c|c|c|c|c|}
\hline
\multicolumn{9}{|c|}{\textbf{\Large Pose Prediction Error}}   \\ \hline
                                                                                 & \multicolumn{2}{c|}{\textbf{\begin{tabular}[c]{@{}c@{}}\large Tap Data\end{tabular}}} & \multicolumn{2}{c|}{\textbf{\begin{tabular}[c]{@{}c@{}}\large Sheared Data\end{tabular}}} & \multicolumn{2}{c|}{\textbf{\begin{tabular}[c]{@{}c@{}}\large Unsheared \large Data\\ \large (Supervised)\end{tabular}}} & \multicolumn{2}{c|}{\textbf{\begin{tabular}[c]{@{}c@{}}\large Unsheared \large Data\\ \large (Semi-Supervised)\end{tabular}}} \\ \hline
\textbf{\begin{tabular}[c]{@{}c@{}}\Large horizontal, \Large$\tau_{x}$\\ \Large(mm)\end{tabular}}   & \multicolumn{2}{c|}{\Large 0.43}                                                                    & \multicolumn{2}{c|}{\Large 2.72}                                                                          & \multicolumn{2}{c|}{\Large 0.64}                                                                        & \multicolumn{2}{c|}{\Large 0.71}                                                                           \\ \hline
\textbf{\begin{tabular}[c]{@{}c@{}}\Large yaw, \Large$\theta n_{Z}$\\ \Large(degrees)\end{tabular}} & \multicolumn{2}{c|}{\Large 2.13}                                                                    & \multicolumn{2}{c|}{\Large 22.20}                                                                         & \multicolumn{2}{c|}{\Large 4.38}                                                                        & \multicolumn{2}{c|}{\Large 4.81}                                                                           \\ \hline
\end{tabular}%
}
\vspace{-1em}
\end{table}

\subsection{Local Contact Geometry Reconstruction}
\begin{figure*}[t]
  \centering
  \includegraphics[width=0.95\textwidth]{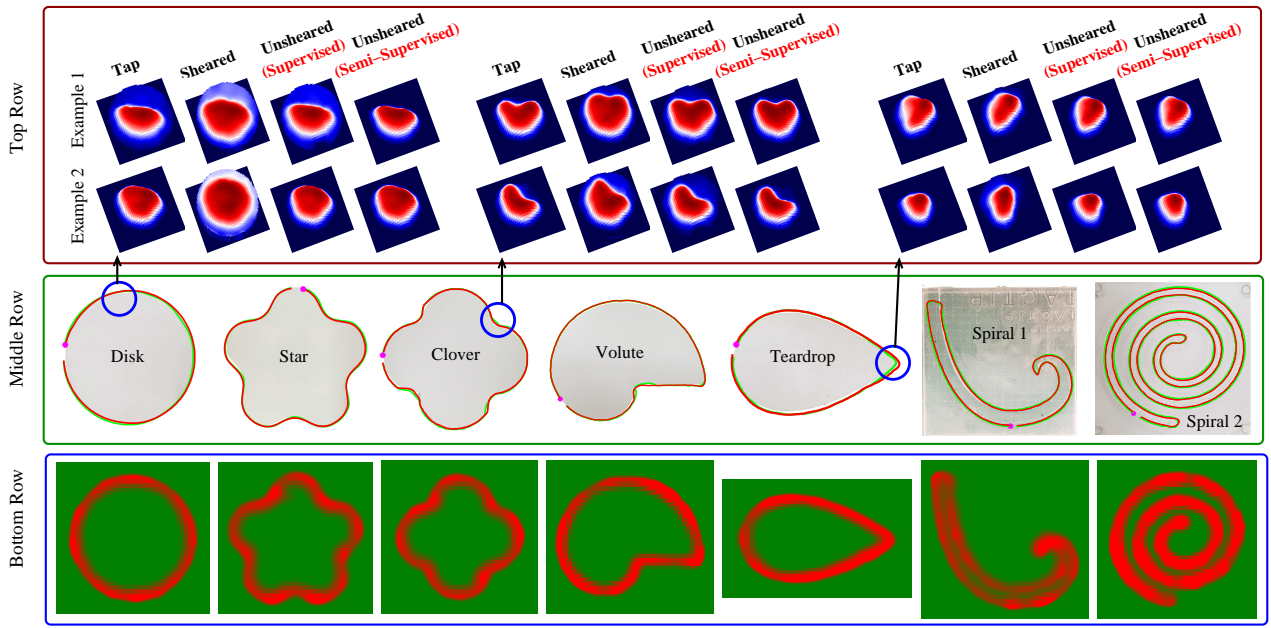}
  \vspace{-0.5em}
  \caption{\textbf{Contact geometry reconstructions, contour following and full object shape reconstructions.} Top row: Local contact geometry reconstruction: The approximate contact geometry can be recovered from sheared data using the model-generated unsheared images but not from the original sheared data. Middle row: Trajectories, overlaid on the objects, under robust sliding using pose estimation from model-generated unsheared images. Red and green trajectories correspond to semi-supervised and supervised~\cite{Anupam_Disentangle_Super} approaches respectively. Red dot is the starting point of trajectory. For both, contact reconstruction and contour following, supervised~\cite{Anupam_Disentangle_Super} \& semi-supervised approaches achieve comparable performance. Bottom row: Full object reconstructions from combining the unsheared local contact geometry reconstructions along the sliding trajectories.}
  \vspace{-1em}
  \label{fig:results}
\end{figure*}

Motion-induced shear masks the contact geometry requiring removal of global shear for its faithful reconstruction (Fig.~\ref{fig:results}, top row). To reconstruct the approximate indentation field, we used a Voronoi-based method~\cite{Nathan_voronoi} that transforms the inner sensor surface with tactile markers into hexagonal cells, thus tessellating the grid of the markers~\cite[Fig. 4]{Nathan_voronoi}. The change in cell areas from an undeformed reference represent the magnitude of local skin deformation that correlates with the indentation due to contact geometry. 

%For display, 
A 3D surface was fitted to the $(x,y)$ centroid coordinates of the Voronoi cells with cell areas as the corresponding height values (Fig.~\ref{fig:results}, top row). The fitted surface from the test sheared data \textit{$S_{\rm inp}$} is visibly distorted by shear (Fig.~\ref{fig:results}, top row). Once global shear is removed in the model-generated unsheared images \textit{$U_{\rm rec}$}, the fitted surface closely resembles the true representation from the original canonical (non-sheared) data \textit{$U_{\rm GT}$} and the unsheared images obtained via supervised baseline~\cite{Anupam_Disentangle_Super} across multiple stimuli (Fig.~\ref{fig:results}, top row). This again demonstrates the effectiveness of our approach in removing global shear \textit{vis-\'{a}-vis} the supervised baseline~\cite{Anupam_Disentangle_Super}. 

\subsection{2D Continuous Shape Exploration}
We further confirmed the model performance by computing the mean-absolute error (MAE) between the predicted pose (lateral position $\tau_{x}$ and in-plane rotation $\theta n_{z}$), from test set model-generated unsheared images \textit{$U_{\rm rec}$}, sheared images \textit{$S_{\rm inp}$} and canonical (non-sheared) images \textit{$U_{\rm GT}$}, and target pose using a pose prediction network \textit{PoseNet} trained on canonical (non-sheared) images previously used in~\cite{Anupam_Disentangle_Super}. In comparison, the prediction error with \textit{$S_{\rm inp}$} was over 4-fold higher than with \textit{$U_{\rm rec}$} (Table~\ref{tab:poseerr}), showing the successful removal of global shear. Our results also show that the model-generated unsheared images from the present model and from the fully supervised baseline~\cite{Anupam_Disentangle_Super} have similar prediction errors (Table~\ref{tab:poseerr}). 

To test the real-time performance and model generalizability to novel situations, we used the \textit{PoseNet}~\cite{Anupam_Disentangle_Super} for sliding exploration of various planar shapes (Fig.~\ref{fig:results}, second row). This testing included four novel shapes absent from training, including two acrylic shapes (spirals 1 and 2, Fig.~\ref{fig:results}, second row) with distinct friction to the 3D-printed stimuli. In addition, for all test shapes, we tested the performance by varying the location of initial contact, depth of indentation, exploration speed, and also the relative location between the object and the sensor tip. The robot successfully traced contours around all test shapes (Fig.~\ref{fig:results}, second row, red) by a combination of model-generated unsheared images, \textit{PoseNet} for prediction of pose parameters and a simple servo control policy~\cite{Anupam_Disentangle_Super}, demonstrating successful removal of global shear. Thus, our model generalizes well to novel experimental conditions. Moreover, comparison of the contours using the current approach (red) and supervised baseline (green)~\cite{Anupam_Disentangle_Super} show near-identical performance (Fig.~\ref{fig:results}, second row).  
 
\subsection{2D Object Reconstruction}
As a final test, we demonstrate our model's utility in the removal of global shear from sheared images by reconstruction of full object shape. This required: 1) faithful contact-geometry reconstruction and 2) successful contour tracing, with both tasks adversely impacted by motion-induced global shear. To obtain full object reconstructions, we first fused together the contact information extracted from unsheared images of sliding contacts recorded during shape exploration and later interpolated them on a rectangular grid (Fig.~\ref{fig:results}, third row) to obtain smooth object reconstructions from discreet sliding contacts. These results demonstrate not only the successful removal of global shear but also the effectiveness of learning the sheared-unsheared mapping once, for later reuse on multiple downstream tasks. The faithful object reconstructions with successful shape exploration across multiple shapes and experimental conditions discussed in section IV.D also show that our model generalizes to novel situations (Fig.~\ref{fig:results}, second and third row).

\section{Discussion}
This work proposed a semi-supervised approach that preserved the sensor deformations due to stimulus contact geometry while removing the shear-distortion caused by motion-dependent shear. We showed that the proposed approach achieves comparable performance to its fully supervised counterpart~\cite{Anupam_Disentangle_Super} with an order-of-magnitude less annotated data, simplifying data collection considerably.

We validated our approach, similar to~\cite{Anupam_Disentangle_Super} by: 1) demonstrating a good match (92.5\%) between the model-generated unsheared images (from sliding contacts) and their non-sheared counterpart taken from vertical tapping, using the structural similarity index measure; 2) reconstructing an approximate stimulus contact geometry from model-generated unsheared images that was previously masked by sliding-induced shear (Fig.~\ref{fig:results}, top row); 3) predicting local object pose from a pose prediction network~\cite{Anupam_Disentangle_Super} trained on non-sheared (tapping) data (Table~\ref{tab:poseerr}). This pose prediction was then used for sliding exploration of multiple planar objects (Fig.~\ref{fig:results}, second row), and so result 4) combined 2) and 3) to reconstruct full object shapes for several planar objects (Fig.~\ref{fig:results}, third row).

%Self-supervision with tactile data differs from computer vision ~\cite{Afros_SplitBrainAE, Noroozi_Jigsaw, Afros_Inpainting} as it lacks attributes like color or texture as well as global data interaction of natural image data. In addition, reliant on tactile data alone eliminated techniques that correlate data across multiple sensory modalities to generate training signals without external supervision~\cite{Owens_AudioVisual_SemiSuper, Raia_Touch_SemiSuper, Lee_Touch_SemiSuper}. In presence of these limitations, we achieved self-supervision by first disentangling sliding data into its contact-only and shear components. Following this, the shear code is recombined with paired and novel canonical (tap) samples to generate training signals via \textit{consistency loss} (see Methods).

One limitation of our study is that it considered only planar objects. We expect our approach will extend to 3D objects, like recent work that has successfully implemented tactile servoing to slide over complex 3D surfaces (3 pose components) and edges (5 pose components)~\cite{Nathan_CNN_Sliding1, Nathan_CNN_Sliding2}. An extension would be to sliding shape exploration and full 3D object reconstruction, which are open problem under active investigation~\cite{suresh2020,bauza2019}. This would open interesting avenues for future research, such as: 1) the fusion of visual and tactile sensory modalities for robust object representations and 2) tactile object recognition by exploiting prior visual experience of the object. An additional limitation is the requirement of annotated data, although much less than the fully supervised counterpart~\cite{Anupam_Disentangle_Super}. An unsolved problem for future work is to completely eliminate external supervision. Finally, our training was limited to translational shear even though servo control introduces rotational shear from the angular motion while sliding over objects. That the methods worked well both for sliding exploration and contact geometry reconstruction in the absence of trying to compensate for rotational shear shows that some shear removal carries over from translation to rotational shear.  Our expectation is that bringing rotational shear into the training will become important when extending the methods to 3D objects.

Overall, we expect that our methods should apply to various other exploration and manipulation tasks using soft optical tactile sensors adversely affected by motion-induced shear.

\end{document}